\documentclass{eptcs}
\providecommand{\event}{ADG 2023} 

\usepackage{proof}
\usepackage{float}
\usepackage{amsmath}
\usepackage{fancyvrb}
\usepackage{fvextra}
\usepackage{xcolor}
\usepackage{url}
\usepackage{xurl}
\usepackage{dirtytalk}
\usepackage{xspace}
\usepackage{csquotes}
\usepackage{tikz}
\usepackage{listings}

\newcommand{\Assumption}{\textsc{Assumption}\xspace}
\newcommand{\MP}{\textsc{MP}\xspace}

\newcommand{\FirstCase}{\textsc{FirstCase}\xspace}
\newcommand{\SecondCase}{\textsc{SecondCase}\xspace}
\newcommand{\QEDbyAssumption}{\textsc{QEDbyAssumption}\xspace}
\newcommand{\QEDbyCases}{\textsc{QEDbyCases}\xspace}
\newcommand{\QEDbyEFQ}{\textsc{QEDbyEFQ}\xspace}

\newcommand{\StepKind}{\texttt{StepKind}\xspace}
\newcommand{\Contents}{\texttt{Contents}\xspace}
\newcommand{\ContentsPredicate}{\texttt{ContentsPredicate}\xspace}
\newcommand{\ContentsArgument}{\texttt{ContentsArgument}\xspace}
\newcommand{\Nesting}{\texttt{Nesting}\xspace}
\newcommand{\AxiomApplied}{\texttt{AxiomApplied}\xspace}
\newcommand{\From}{\texttt{From}\xspace}
\newcommand{\Instantiation}{\texttt{Instantiation}\xspace}
\newcommand{\Cases}{\texttt{Cases}\xspace}

\newcommand{\Goal}{\texttt{Goal}\xspace}

\allowdisplaybreaks

\usepackage[breakable,skins,theorems]{tcolorbox}
\tcbuselibrary{skins,breakable}
\tcbset{highlight math style={enhanced,breakable,colframe=red,colback=white,arc=0pt,boxrule=1pt}}

\usepackage{graphicx}
\usepackage{subcaption}
\usepackage{pgfplots}
\pgfplotsset{compat=1.15}
\usepackage{mathrsfs}
\usetikzlibrary{arrows}
\usepackage[strings]{underscore}

\newtheorem{definition}{Definition}
\newtheorem{example}{Example}

\definecolor{uuuuuu}{rgb}{0.26666666666666666,0.26666666666666666,0.26666666666666666}
\definecolor{zzttqq}{rgb}{0.1,0.2,0.3}
\definecolor{ududff}{rgb}{0.30196078431372547,0.30196078431372547,1}
\definecolor{xdxdff}{rgb}{0.49019607843137253,0.49019607843137253,1}
\definecolor{qqwuqq}{rgb}{0,0.39215686274509803,0}
\definecolor{qqqqff}{rgb}{0,0.39215686274509803,0}

\newcounter{proofstepnum}

\newenvironment{changemargin}[3]{%
\begin{list}{}{%
\setlength{\topsep}{0pt}%
\dimen0=10pt\relax
\setlength{\leftmargin}{#2\dimen0}%
\setlength{\rightmargin}{#3}%
\setlength{\listparindent}{\parindent}%
\setlength{\itemindent}{\parindent}%
\setlength{\parsep}{\parskip}%
}%
\item[#1.]}
{\end{list}}
\def\proofstep#1#2{
\addtocounter{proofstepnum}{1}
\begin{changemargin}{\theproofstepnum}{#1}{0pt}
#2
\end{changemargin}
}
\newenvironment{changemargin1}[2]{%
\begin{list}{}{%
\setlength{\topsep}{0pt}%
\dimen0=3pt\relax
\setlength{\leftmargin}{#1\dimen0}%
\setlength{\rightmargin}{#2}%
\setlength{\listparindent}{\parindent}%
\setlength{\itemindent}{\parindent}%
\setlength{\parsep}{\parskip}%
}%
\item[]}
{\end{list}
}

\author{Salwa Tabet Gonzalez \and Predrag Jani\v{c}i\'c \and Julien Narboux}

\title{Automated Completion of Statements and Proofs in
Synthetic \\ Geometry: an Approach based on Constraint Solving}

\author{
Salwa Tabet Gonzalez
    \institute{UMR 7357 CNRS \\ University of Strasbourg \\
    P\^{o}le API, Bd S\'ebastien Brant \\ BP 10413 \\ 67412 Illkirch, France}
    \email{tabetgonzalez@unistra.fr}
\and
Predrag Jani\v{c}i\'c
\institute{Department for Computer Science \\ Faculty of Mathematics \\
University of Belgrade \\ Studentski trg 16 \\ 11000 Belgrade, Serbia}
\email{janicic@matf.bg.ac.rs}
\and
Julien Narboux
\institute{UMR 7357 CNRS \\ University of Strasbourg \\
P\^{o}le API, Bd S\'ebastien Brant \\ BP 10413 \\ 67412 Illkirch, France}
\email{narboux@unistra.fr}
}

\def\titlerunning{Automated Completion of Statements and Proofs in Synthetic
Geometry}
\def\authorrunning{Gonzalez \& Janičić \& Narboux}

\begin{document}

\maketitle

\begin{abstract}
Conjecturing and theorem proving are activities at the center
    of mathematical practice and are difficult to separate.
    In this paper, we propose a framework for completing incomplete
    conjectures and incomplete proofs. The framework can turn a conjecture with missing assumptions and with an under-specified
    goal into a proper theorem.
    Also, the proposed framework can help in completing a proof sketch
    into a human-readable and machine-checkable proof.
    Our approach is focused on synthetic geometry, and uses coherent
    logic and constraint solving.
    The proposed approach is uniform for all three kinds of tasks,
    flexible and, to our knowledge, unique such approach.
\end{abstract}

\definecolor{wrwrwr}{rgb}{0.3803921568627451,0.3803921568627451,0.3803921568627451}
\definecolor{rvwvcq}{rgb}{0.08235294117647059,0.396078431372549,0.7529411764705882}

\section{Introduction}
\label{sec:intro}

Automated theorem provers take as input the formal statement of a conjecture in a theory described by axioms and lemmas, and try to generate a proof or a counter-example for this conjecture.
In the field of geometry, several efficient automated theorem proving approaches have been developed, including
algebraic ones such as Wu's method, Gröbner bases method, and semi-synthetic methods such as the area method.
In these approaches, typically, the conjecture and the axioms
being considered are fixed.
However, in mathematical practice, in the context of education and
also in mathematical research, the conjecturing and 
proving activities are not separated but interleaved. 
The practitioner may try to prove a statement which is valid only assuming 
some implicit or unknown assumptions, while the list of lemmas and theorem which can be used may not be complete.
 In education, for some kind of exercises, a precise formulation of the statement to be proved is also left to the student, with questions such as:
 \say{What is the nature of the quadrilateral $ABCD$?}.
Hence, the conjecture can contain unknown assumptions called \emph{abducts}, and the goal may be not completely specified.
One may also ask for a proof using a particular theorem or an intermediate fact, i.e., a proof partially specified using constraints 
specifying some proof steps.

In this paper, we consider the problems of (simultaneously) completing
(a) the assumptions of the conjecture;
(b) the goal of the conjecture;
(c) a proof sketch for the conjecture.
The completion process should lead to a proof that is both machine-checkable and human-readable. Because we aim at producing intelligible and readable proofs,
with a similar level of granularity as paper-and-pencil proofs,
our approach is logic-based, uses a fragment of first-order logic
called coherent logic, and is focused on synthetic geometry (in contrast
to algebraic methods).
Our approach for dealing with partial conjectures and partial proofs
is implemented as an extension of the automated theorem prover
Larus developed previously~\cite{janicic_theorem_2022}.
The approach is uniform for all three kinds of completion tasks, flexible
and, to our knowledge, unique such approach.

\begin{figure}[t!]
\begin{center}
\begin{tikzpicture}
\clip (0,0) rectangle (9.000000,4.000000);
{\footnotesize

\definecolor{r200g200b250}{rgb}{0.784314,0.784314,0.980392}%
\color{r200g200b250}%

\fill (0.600000,3.500000) -- (0.300000,0.300000) -- (1.200000,1.500000);%

\fill (0.300000,0.300000) -- (1.200000,1.500000) -- (2.500000,1.500000);%

\definecolor{r200g220b250}{rgb}{0.784314,0.862745,0.980392}%
\color{r200g220b250}%

\fill (0.450000,1.900000) -- (1.400000,0.900000) -- (1.850000,1.500000);%

\fill (0.450000,1.900000) -- (0.900000,2.500000) -- (1.850000,1.500000);%

\definecolor{r0g0b250}{rgb}{0.000000,0.000000,0.980392}%
\color{r0g0b250}%

\draw [line width=0.048cm] (0.477550,1.871000) -- (1.372450,0.929000);%

\draw [line width=0.048cm] (1.424000,0.932000) -- (1.826000,1.468000);%

\draw [line width=0.048cm] (1.822450,1.529000) -- (0.927550,2.471000);%

\draw [line width=0.048cm] (0.876000,2.468000) -- (0.474000,1.932000);%

\definecolor{r0g0b0}{rgb}{0.000000,0.000000,0.000000}%
\color{r0g0b0}%

\draw [line width=0.048cm] (0.596266,3.460175) -- (0.453734,1.939825);%
\draw [line width=0.048cm] (0.446266,1.860175) -- (0.303734,0.339825);%

\draw [line width=0.048cm] (0.335116,0.319154) -- (1.364884,0.880846);%
\draw [line width=0.048cm] (1.435116,0.919154) -- (2.464884,1.480846);%

\draw [line width=0.048cm] (2.460000,1.500000) -- (1.890000,1.500000);%
\draw [line width=0.048cm] (1.810000,1.500000) -- (1.240000,1.500000);%

\draw [line width=0.048cm] (1.188506,1.538313) -- (0.911494,2.461687);%
\draw [line width=0.048cm] (0.888506,2.538313) -- (0.611494,3.461687);%

\draw [line width=0.048cm] (0.600000,3.500000) circle (0.040000);%
\draw (0.600000,3.500000) node [anchor=south] { $A$ };%

\draw [line width=0.048cm] (0.300000,0.300000) circle (0.040000);%
\draw (0.330000,0.330000) node [anchor=north east] { $B$ };%

\draw [line width=0.048cm] (2.500000,1.500000) circle (0.040000);%
\draw (2.500000,1.500000) node [anchor=north] { $C$ };%

\draw [line width=0.048cm] (1.200000,1.500000) circle (0.040000);%
\draw (1.170000,1.470000) node [anchor=south west] { $D$ };%

\draw [line width=0.048cm] (0.450000,1.900000) circle (0.040000);%
\draw (0.450000,1.900000) node [anchor=east] { $E$ };%

\draw [line width=0.048cm] (1.400000,0.900000) circle (0.040000);%
\draw (1.400000,0.900000) node [anchor=north] { $F$ };%

\draw [line width=0.048cm] (1.850000,1.500000) circle (0.040000);%
\draw (1.820000,1.470000) node [anchor=south west] { $G$ };%

\draw [line width=0.048cm] (0.900000,2.500000) circle (0.040000);%
\draw (0.870000,2.470000) node [anchor=south west] { $H$ };%

\definecolor{r200g200b250}{rgb}{0.784314,0.784314,0.980392}%
\color{r200g200b250}%

\fill (3.900000,3.500000) -- (3.000000,1.000000) -- (5.600000,3.100000);%

\fill (3.000000,1.000000) -- (5.600000,3.100000) -- (5.100000,0.300000);%

\definecolor{r200g220b250}{rgb}{0.784314,0.862745,0.980392}%
\color{r200g220b250}%

\fill (3.450000,2.250000) -- (4.050000,0.650000) -- (5.350000,1.700000);%

\fill (3.450000,2.250000) -- (4.820000,3.565000) -- (5.350000,1.700000);%

\definecolor{r0g0b250}{rgb}{0.000000,0.000000,0.980392}%
\color{r0g0b250}%

\draw [line width=0.048cm] (3.464045,2.212547) -- (4.035955,0.687453);%

\draw [line width=0.048cm] (4.081118,0.675133) -- (5.318882,1.674867);%

\draw [line width=0.048cm] (5.339066,1.738476) -- (4.830934,3.526524);%

\draw [line width=0.048cm] (4.791142,3.537301) -- (3.478858,2.277699);%

\definecolor{r0g0b0}{rgb}{0.000000,0.000000,0.000000}%
\color{r0g0b0}%

\draw [line width=0.048cm] (3.886451,3.462365) -- (3.463549,2.287635);%
\draw [line width=0.048cm] (3.436451,2.212365) -- (3.013549,1.037635);%

\draw [line width=0.048cm] (3.037947,0.987351) -- (4.012053,0.662649);%
\draw [line width=0.048cm] (4.087947,0.637351) -- (5.062053,0.312649);%

\draw [line width=0.048cm] (5.107032,0.339377) -- (5.342968,1.660623);%
\draw [line width=0.048cm] (5.357032,1.739377) -- (5.592968,3.060623);%

\draw [line width=0.048cm] (5.561063,3.109162) -- (3.938937,3.490838);%

\draw [line width=0.048cm] (3.900000,3.500000) circle (0.040000);%
\draw (3.900000,3.500000) node [anchor=south] { $A$ };%

\draw [line width=0.048cm] (3.000000,1.000000) circle (0.040000);%
\draw (3.030000,1.030000) node [anchor=north east] { $B$ };%

\draw [line width=0.048cm] (5.100000,0.300000) circle (0.040000);%
\draw (5.100000,0.300000) node [anchor=north] { $C$ };%

\draw [line width=0.048cm] (5.600000,3.100000) circle (0.040000);%
\draw (5.570000,3.070000) node [anchor=south west] { $D$ };%

\draw [line width=0.048cm] (3.450000,2.250000) circle (0.040000);%
\draw (3.450000,2.250000) node [anchor=east] { $E$ };%

\draw [line width=0.048cm] (4.050000,0.650000) circle (0.040000);%
\draw (4.050000,0.650000) node [anchor=north] { $F$ };%

\draw [line width=0.048cm] (5.350000,1.700000) circle (0.040000);%
\draw (5.350000,1.700000) node [anchor=west] { $G$ };%

\draw [line width=0.048cm] (4.820000,3.565000) circle (0.040000);%
\draw (4.790000,3.535000) node [anchor=south west] { $H$ };%

\definecolor{r200g200b250}{rgb}{0.784314,0.784314,0.980392}%
\color{r200g200b250}%

\fill (7.800000,3.500000) -- (6.100000,2.300000) -- (8.700000,2.300000);%

\fill (6.100000,2.300000) -- (8.700000,2.300000) -- (8.000000,0.300000);%

\definecolor{r200g220b250}{rgb}{0.784314,0.862745,0.980392}%
\color{r200g220b250}%

\fill (6.950000,2.900000) -- (7.050000,1.300000) -- (8.350000,1.300000);%

\fill (6.950000,2.900000) -- (8.250000,2.900000) -- (8.350000,1.300000);%

\definecolor{r0g0b250}{rgb}{0.000000,0.000000,0.980392}%
\color{r0g0b250}%

\draw [line width=0.048cm] (6.952495,2.860078) -- (7.047505,1.339922);%

\draw [line width=0.048cm] (7.090000,1.300000) -- (8.310000,1.300000);%

\draw [line width=0.048cm] (8.347505,1.339922) -- (8.252495,2.860078);%

\draw [line width=0.048cm] (8.210000,2.900000) -- (6.990000,2.900000);%

\definecolor{r0g0b0}{rgb}{0.000000,0.000000,0.000000}%
\color{r0g0b0}%

\draw [line width=0.048cm] (7.767321,3.476933) -- (6.982679,2.923067);%
\draw [line width=0.048cm] (6.917321,2.876933) -- (6.132679,2.323067);%

\draw [line width=0.048cm] (6.127550,2.271000) -- (7.022450,1.329000);%
\draw [line width=0.048cm] (7.077550,1.271000) -- (7.972450,0.329000);%

\draw [line width=0.048cm] (8.013214,0.337754) -- (8.336786,1.262246);%
\draw [line width=0.048cm] (8.363214,1.337754) -- (8.686786,2.262246);%

\draw [line width=0.048cm] (8.676000,2.332000) -- (8.274000,2.868000);%
\draw [line width=0.048cm] (8.226000,2.932000) -- (7.824000,3.468000);%

\draw [line width=0.048cm] (7.800000,3.500000) circle (0.040000);%
\draw (7.800000,3.500000) node [anchor=south] { $A$ };%

\draw [line width=0.048cm] (6.100000,2.300000) circle (0.040000);%
\draw (6.130000,2.330000) node [anchor=north east] { $B$ };%

\draw [line width=0.048cm] (8.000000,0.300000) circle (0.040000);%
\draw (8.000000,0.300000) node [anchor=north] { $C$ };%

\draw [line width=0.048cm] (8.700000,2.300000) circle (0.040000);%
\draw (8.670000,2.270000) node [anchor=south west] { $D$ };%

\draw [line width=0.048cm] (6.950000,2.900000) circle (0.040000);%
\draw (6.980000,2.870000) node [anchor=south east] { $E$ };%

\draw [line width=0.048cm] (7.050000,1.300000) circle (0.040000);%
\draw (7.050000,1.300000) node [anchor=north] { $F$ };%

\draw [line width=0.048cm] (8.350000,1.300000) circle (0.040000);%
\draw (8.350000,1.300000) node [anchor=west] { $G$ };%

\draw [line width=0.048cm] (8.250000,2.900000) circle (0.040000);%
\draw (8.220000,2.870000) node [anchor=south west] { $H$ };%
\color{black}
}
\end{tikzpicture}
\end{center}
\caption{Illustrations for five problems related to Varignon's
theorem, respectively: Problem 1; Problem 2; Problem 3.}
\label{fig:varignon}
\end{figure}
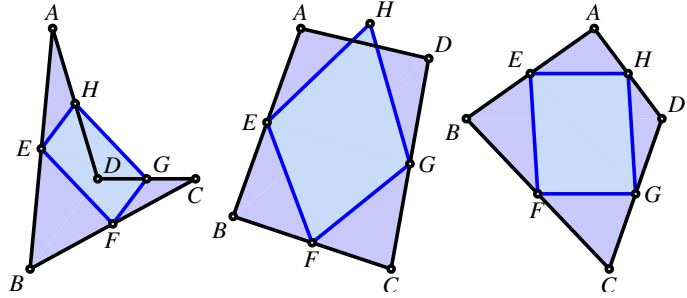

We list five high-school level synthetic geometry problems related to Varignon's theorem (Figure \ref{fig:varignon}), that we will try to solve using our approach.

\begin{description}
\item[Problem 1 (Fully specified statement)] Consider a quadrilateral $ABCD$, let $E$, $F$, $G$ and $H$ be the midpoints of $AB$, $BC$,
$CD$, $DA$ respectively. Prove that the quadrilateral $EFGH$ is
a parallelogram (assuming that there are no two sides that are aligned).
\item[Problem 2 (First inverse problem)] Consider a quadrilateral $ABCD$, let $E$, $F$, 
and $G$ be the midpoints of $AB$, $BC$ and $CD$ respectively.
Let $H$ be a point. 
Under which assumption is the quadrilateral $EFGH$ a parallelogram?
\item[Problem 3 (Second inverse problem)]
Consider a quadrilateral $ABCD$, let $E$, $F$, $G$ and $H$ 
be the midpoints of $AB$, $BC$, $CD$, $DA$ respectively.
Under which assumption is the quadrilateral $EFGH$ a rectangle? 
\item[Problem 4 (Partially specified goal)]
Consider a quadrilateral $ABCD$, let $E$, $F$, $G$ and $H$ 
be the midpoints of $AB$, $BC$, $CD$, $DA$ respectively.
What is the nature of the quadrilateral $EFGH$? 
\item[Problem 5 (Partially specified proof)] 
Consider a quadrilateral $ABCD$, let $E$, $F$, $G$ and $H$ 
be the midpoints of $AB$, $BC$, $CD$, $DA$ respectively. We have
that $EG = FH$.
Prove that $EFGH$ is a rectangle using the axiom \say{If the diagonals of 
a parallelogram are congruent, then it's a rectangle}.
\end{description}

\noindent
The above examples are inspired by exercises given in a teacher training session.
A more detailed discussion about how these examples can be used in a  didactic context, issues related to the formalization can be found in~\cite{durand-guerrier_examining_2012,narboux_combining_2021}

\section{Background}
\label{sec:background}

This section provides some necessary background information on
a fragment of first-order logic called coherent logic that
our approach uses. There are several automated provers for
coherent logic, including Larus, which is based on \say{theorem proving
as constraint solver} paradigm.

\subsection{Coherent Logic}
\label{subsec:cl}

A formula of first-order logic is said to be {\em coherent}
if it has the following form:
$$A_0(\vec{x}) \wedge \ldots \wedge A_{n-1}(\vec{x}) \Rightarrow
\exists \vec{y} ( B_0(\vec{x},\vec{y}) \vee \ldots \vee \; B_{m-1}(\vec{x},\vec{y}))$$
\noindent
where universal closure is assumed, and where
$\vec{x}$ denotes a sequence of variables $x_0, x_1, \ldots, x_{k-1}$;
$A_i$ (for $0 \leq i \leq n-1$) denotes an atomic
formula (involving zero or more variables from $\vec{x}$);
$\vec{y}$ denotes a sequence of variables
$y_0, y_1, \ldots, y_{l-1}$; $B_j$ (for $0 \leq j \leq m-1$)
denotes a conjunction of atomic formulae (involving zero or more
of the variables from $\vec{x}$ and $\vec{y}$) \cite{janicic_theorem_2022}.
If there are no formulae $A_i$, then the
left-hand side of the implication is assumed to be $\top$. If
there are no formulae $B_j$, then the right-hand side of the implication
is assumed to be $\bot$. There are no function
symbols with arity greater than zero. Coherent formulae do not
involve   the negation connective. A coherent theory is a set of sentences,
axiomatized by coherent formulae, and closed under derivability.
A number of theories and theorems can be formulated directly and
simply in coherent logic (CL). 
In addition, any first-order theory can be translated into CL,
possibly with additional predicate symbols
\cite{dyckhoff_geometrization_2015,polonsky_proofs_2011}.
Synthetic geometry can be expressed easily using CL.
For example, the central part of axioms system of Euclid (as formalized by Beeson et al.~\cite{beeson_proof-checking_2019}), or Hilbert (as formalized by Braun et al.~\cite{braun_tarski_2012}), or Tarski~\cite{schwabhauser_metamathematische_1983}
can be expressed in first-order logic without function symbols, and the
axioms are mostly in CL form.

Translation of FOL formulae into CL involves elimination of the
negation connectives: negations can be kept in place and new
predicates symbols for corresponding sub-formula have to be
introduced, or negations can be pushed down to atomic formulae~\cite{polonsky_proofs_2011}.
In the latter case, for every predicate symbol $R$ (that appears
in negated form), a new symbol $\overline{R}$ is introduced that
stands for $\neg R$, and the following axioms are introduced:
$\forall \vec{x} (R(\vec{x}) \wedge \overline{R}(\vec{x}) \Rightarrow \bot)$,
$\forall \vec{x} (R(\vec{x}) \vee \overline{R}(\vec{x}))$.

In contrast to resolution-based theorem proving, in forward
reasoning for CL, the conjecture being proved is kept unchanged
and proved without using refutation, Skolemization and clausal
form. Thanks to this, CL is suitable for producing human-readable
synthetic proofs and also machine verifiable proofs  \cite{bezem_automating_2005,dyckhoff_geometrization_2015}.
The problem of provability in CL is semi-decidable. CL admits
a simple proof system, a sequent-based variant is as follows~\cite{stojanovic_vernacular_2014}:

\begin{tcolorbox}[breakable]
$$\infer[ \mathrm{MP}]
{\Gamma, ax, A_0(\vec{a}), \ldots , A_{n-1}(\vec{a}) \vdash P}
{\Gamma, ax, A_0(\vec{a}), \ldots , A_{n-1}(\vec{a}), \underline{B_0(\vec{a},\vec{b})
\vee \ldots \vee B_{m-1}(\vec{a},\vec{b})} \vdash P}
$$

$$\infer[  \mathrm{QEDcs \; (case \; split)}]
{\Gamma,B_0(\vec{c}) \vee \ldots \vee B_{m-1}(\vec{c}) \vdash P}
{\Gamma,\underline{B_0(\vec{c})} \vdash P   \,\,\,\,\,\,\,    \ldots \,\,\,\,\,\,\,  \Gamma, \underline{B_{m-1}(\vec{c})} \vdash P}
$$

$$\infer[  \mathrm{QEDas \; (assumption)} ]
{\Gamma, \underline{B_i(\vec{a},\vec{b})} \vdash \exists \vec{y} ( B_0(\vec{a},\vec{y}) \vee \ldots \vee \; B_{m-1}(\vec{a},\vec{y}))}
{}
$$

$$\infer[  \mathrm{QEDefq \; (ex \; falso \; quodlibet)} ]
{\Gamma,\bot \vdash P}
{}
$$
\end{tcolorbox}

\noindent
In the rules given above, it is assumed: $ax$ is a formula
$A_0(\vec{x}) \wedge \ldots \wedge A_{n-1}(\vec{x}) \Rightarrow
\exists \vec{y} ( B_0(\vec{x},\vec{y}) \vee \ldots \vee \; B_{m-1}(\vec{x},\vec{y}))$;\footnote{Notice
the hidden link between the formulae $B_i(\vec{a},\vec{b})$
from the rule $\mathrm{MP}$ and the formula $ax$: the formulae
$B_i(\vec{a},\vec{b})$ from the rule
are instances of the formulae $B_i(\vec{x},\vec{y})$ from $ax$.}
$\vec{a}$, $\vec{b}$, $\vec{c}$ denote sequences of constants
(possibly of length zero); in the rule $\mathrm{MP}$
({\em extended \textit{modus ponens}}), $\vec{b}$ are fresh constants;
$\vec{x}$ and $\vec{y}$ denote sequences of variables (possibly
of length zero); $A_i(\vec{x})$ (respectively $B_i(\vec{x},\vec{y})$)
have no free variables other than from $\vec{x}$ (respectively $\vec{x}$ and $\vec{y}$); $A_i(\vec{a})$ are ground atomic formulae; $B_i(\vec{a},\vec{b})$
and $B_i(\vec{c})$ are conjunctions of ground atomic formulae;
$\underline{\Phi}$ denotes the list of conjuncts in $\Phi$ if
$\Phi$ is conjunction, and otherwise $\Phi$ itself. In the proving
process, the rules are read from bottom to top, i.e., by a rule
application one gets the contents (new sub-goals) above the
line.

For a set of coherent axioms $\mathit{AX}$ and the statement
$A_0(\vec{x}) \wedge \ldots \wedge A_{n-1}(\vec{x}) \Rightarrow
\exists \vec{y} ( B_0(\vec{x},\vec{y}) \vee \ldots \vee \; B_{m-1}(\vec{x},\vec{y}))$
to be proved, within the above proof system one has to derive
the following sequent (where $\vec{a}$ denotes a sequence of new
symbols of constants):
$\mathit{AX}, A_0(\vec{a}), \ldots, A_{n-1}(\vec{a}) \vdash
\exists \vec{y} (B_0(\vec{a},\vec{y}) \vee \ldots \vee \; B_{m-1}(\vec{a},\vec{y}))$.

Notice that, in the above proof system, case split may
occur only at the end of a (sub)proof. However, it is not
a substantial restriction: any proof with unrestricted use
of case split can be transformed to such form.


\subsection{Theorem Proving as Constraint Solving and the Larus System}
\label{subsec:larus}

 \enquote{Theorem proving as constraint solving} is a paradigm
for automated theorem proving recently proposed \cite{janicic_theorem_2022}.
In contrast to common automated theorem proving approaches,
in which the search space is a set of some formulae and what
is sought is again a (goal) formula, this new approach is
based on searching for a proof (of a given length) as a whole.
Namely, a proof of a formula in a fixed logical setting can be
encoded as a sequence of natural numbers obeying some constraints.
A suitable solver can find such a sequence and from that
sequence a sought proof can be reconstructed.
This approach is implemented in C++, within an open-source
prover Larus,\footnote{\url{https://github.com/janicicpredrag/Larus}}
specialized in proofs in coherent logic and using SAT, SMT, and
CSP solvers for solving sets of constraints. Larus can generate readable, human understandable proofs in natural language and also machine-verifiable proofs for the interactive provers Coq, Isabelle,
and Mizar.

Each CL proof consists of several proof steps, while each of
them has one of the following kinds (with obvious meaning):
\Assumption, \MP, \FirstCase, \SecondCase, \QEDbyCases,
\QEDbyAssumption, \QEDbyEFQ.
The information relevant for \MP steps include:     
\AxiomApplied, \From (the ordinal numbers of proof steps
justifying premises of the axiom applied),
\Instantiation (of the variables in the axiom),
\Contents (the atoms in formula in the proof step), etc.
\Nesting denotes the nesting of the proof
step (the nesting of the first step is 1).

The proof can be represented by a sequence of numbers,
meeting some constraints (that correspond to definitions
of inference steps given in Section \ref{subsec:cl}).
For instance, if the proof step $s$ is of the kind \QEDbyEFQ,
then the following conditions must hold
(given almost in verbatim as in our C++ code):\footnote{
The corresponding C++ implementation is an improved version
of the implementation presented earlier \cite{janicic_theorem_2022}.}

    \begin{enumerate}
        \item \StepKind$(s) = $ \QEDbyEFQ;
        \item $s>0$;
        \item \Contents$(s-1)(0) = \bot$;
        \item \Goal$(s)$;
        \item \Nesting$(s) = $ \Nesting$(s-1)$.
    \end{enumerate}

\noindent
The above conditions can be understood in the following way:
if there is a proof of the given conjecture, the proof step $s$
in that proof is of the kind \QEDbyEFQ iff the natural number
\StepKind$(s)$ equals the code for $\QEDbyEFQ$,
$s>0$ (since there must be a previous step),
the contents of the previous proof step is $\bot$,
the contents of the step is the goal itself,
and the nesting of the steps $s-1$ and $s$
is the same.

Each proof step has one of the listed kinds and meet
corresponding conditions. There are also some additional,
global constraints, like that the last proof step has
\Nesting equal 1.

Larus works in the following way. If there is a set of axioms,
a conjecture, and a proof length, a corresponding proof can
be represented as a sequence of natural numbers, still unknown,
so they will be represented by variables $V$. The constraints
that have to be met for each proof step and for the proof
as a whole
can be expressed in terms of these variables $V$. If a solver
can find a model for the constraint, from it the proof in
logical terms can be reconstructed. All constraints involved
are linear constraints over natural numbers. Since linear
arithmetic is decidable, decision procedures for it can
decide, for each input constrains, whether or not it has a
model. For this purpose, Larus can use SAT, SMT, and CSP
solvers. For input, Larus uses axioms and conjectures
stored in a file in TPTP/fof format.

\section{Abducts and Completing Assumptions}
\label{sec:abducts}

There are three major types of logical inference: induction, deduction,
and abduction. The concept of abduction has been introduced by Peirce~\cite{peirce_collected_1932}. In deduction, everything inferred
is necessarily true, while it is not the case with the remaining two types
of inference. Induction tries to infer general rules based on individual
instances. The aim of abduction is to produce additional hypotheses to
explain observed facts. Abduction has a wide spectrum of implicit or
explicit applications -- in everyday life, in education, and in scientific
reasoning, including in building mathematical theories, or in software
verification. One definition of abduct is given below.

\begin{definition}
\label{def:abduction}
Given a theory $T$ and a formula $G$ (the goal to be proved), such that
$T \not \models G$,
an {\em explanations} or {\em abduct} is a formula $A$ meeting conditions: $T, A \models G$ and $T, A \not \models \bot$.
\end{definition}

\noindent
It is clear that some abducts are not \emph{interesting},
so there are often some additional restrictions given.
There is no general agreement about such restrictions, but
two types are most usual: {\em syntactical restrictions} (abducts should
be of a specific syntactical form) and {\em minimality restrictions}
(for any other abduct $A'$, if $T,A \models A'$ then $A \equiv A'$).
It is reasonable to ask that $A$ is not $G$, as it is trivial.
Some authors also add stronger a restriction that $A \not \models G$
(i.e., at least one axiom of $T$ has to be used to prove $G$).

\paragraph{Approaches for Computing Abducts.}
Various algorithms to produce different kind of abducts have been developed \cite{atocha_abductive_2006}.
In \emph{abductive logic programming}, techniques for abductive reasoning are developed in the context of logic programming.
Rules are considered to be Horn clauses~\cite{denecker_abduction_2002}.
According to Russo et al.~\cite{russo_use_2001}, some systems assume that predicate symbols appearing in abducts do not appear in the conclusion of any rule
 and that negation does not appear in the conclusion of any rule. This restriction is not realistic in the context of geometry.
 In our example, we want to accept geometric predicate symbols both in abducts, and in the assumptions and conclusion of theorems.
 Some approaches are based on Robinson's resolution algorithm, extended such that when no more clauses can be produced, the atomic clauses are considered as a potential abduct and consistency if checked~\cite{marquis_extending_1991}.
 There are also approaches developed for the context
 of SMT solving, dealing with decidable theories like linear
 arithmetic~\cite{dillig_explain_2013,reynolds_scalable_2020}

In the context of geometry, some algebraic algorithms can generate additional assumptions for the statement to be true.
For example, Wu's method \cite{wu_decision_1978} can produce non-degeneracy conditions.
Algebraic methods can also be used to generate more general abducts \cite{recio_automatic_1999}.
These methods are more efficient than ours, but
more specific so cannot be used for arbitrary geometric theories.
Also, they cannot generate readable proofs. 
Moreover, expressing algebraic non-degeneracy conditions in simple geometrical terms is not easy and not always possible
\cite{chen_projection_2004}.

\paragraph*{Abduction in Synthetic Euclidean Geometry.}

In this paper, the theory $T$ from Definition \ref{def:abduction}
is a synthetic Euclidean geometry. In this context, automated finding
of proofs allowing abducts may have several applications. For instance,
an automated system may help a student or a researcher who tries to prove
(or formalize) a theorem with a missing assumption. 
Barbosa et al.~have proposed such goal (although not for geometry) in the context of  interactive proof assistants where conjectures are sent to an SMT solver~\cite{barbosa_interactive_2023}.

Non-degeneracy conditions are often overlooked and missing in informal geometry statements.
Abductive reasoning is also a task which can be asked explicitly
to students. The answer expected by the teacher for Problem 2 is
that $H$ should be the midpoint of $AD$.

\paragraph*{Finding Abducts using Larus.}

In this paper, we restrict consideration of abduction only to coherent logic and only to abducts that are conjunctions of ground atomic formulae. Larus was not implemented with
abduction in mind, yet implementation of support for abduction
turned out to be very simple, almost trivial, and took less
than 100 lines of C++ code.
In order to find abducts using Larus, we treat them as a special case
of proof steps, in the main proof branch, just after assumptions. We
have to add constraints on what such an abduct can be:
\begin{enumerate}
 \item the abduct is treated as an assumption;
 \item the nesting of the abduct equals $1$;
 \item the abduct is an atomic formula (no branching);
 \item the predicate symbol is one of the predicate
symbols in the signature;
 \item the arguments are among existing symbols of constants;
 \item the abduct is not the goal itself;
 \item the abduct is not $\bot$.
\end{enumerate}

The given conditions may be written in the following way,
assuming that the abduct is placed in $i$-th step of the
proof:

\begin{enumerate}
\item \StepKind$(i) = $\Assumption
\item \Nesting$(i) = 1$
\item \Cases$(i) = \mathit{false}$
\item \ContentsPredicate$(i,0) < sizeof(Signature)$
\item for each argument $j$ (up to maximal arity):
      \ContentsArgument$(i,0,j) < sizeof(Constants)$
\item \Goal$(i) = \mathit{false}$
\item \ContentsPredicate$(i,0) \; \neq \bot$
\end{enumerate}

\noindent
One can also choose a number of abducts, each leading to the
constraints given above. With such additional constraints
for each abduct (for additional proof steps in specific
positions in a proof sought), with a given set of
axioms and a conjecture, and with a concrete proof length,
we run Larus as usual.
The solving/proving process is the same as without abducts:
the constraint solver finds a way to specify a full proof,
including the abducts, i.e., under-specified assumptions.

In the above list of conditions, the last two do not follow
the basic definition of abduct. Like in some other variants of the
definition,  the abduct may not be equal to the goal
atom because such abducts are trivial. Also, the abduct may not
be equal to $\bot$, since it is inconsistent. It is important to
discard other inconsistent abducts early, so we add one more
restriction: the proof of $T, A \models G$ should not end with
\QEDbyEFQ.
Some constructed abducts may still be inconsistent with
other assumptions, and we use an external, more efficient
automatic theorem prover, Vampire~\cite{kovacs_first-order_2013}, to discard such abducts.

\begin{example}
For the first inverse problem (Problem 2 from Section \ref{sec:intro}),
Larus produces two consistent, symmetric abducts (the proof obtained
with the first abduct is presented in Appendix \ref{annex:varignonp2}):
\begin{itemize}
	\item \say{$H$ is the midpoint of $AD$}
	\item \say{$H$ is the midpoint of $DA$}
\end{itemize}
 \end{example}

\begin{example}
For the second inverse problem (Problem 3 from Section \ref{sec:intro}), Larus produces more than 150 consistent abducts, most of which give degenerate cases, hence are less interesting. Apart from such abducts, we obtained the following abducts and their symmetric variants
(the proof with the first abduct is presented in Appendix \ref{annex:varignonp3}):
\begin{itemize}
	\item \say{the diagonals $HF$ and $EG$ are congruent}
	\item \say{$\angle FGH$ is a right angle}
	\item \say{$\angle EHG$ is a right angle}
	\item \say{$\angle HEF$ is a right angle}
	\item \say{$\angle EFG$ is a right angle}
\end{itemize}
\end{example}

\section{Deducts and Completing Goals}
\label{sec:deducts}

Non-trivial first-order logic theories have infinite
number of theorems.
Approaches based on refutation 
cannot be used with under-specified goals and, hence,
cannot be used for completing them.
In principle, a controlled forward-reasoning (for instance,
based on some kind of breadth-first search) can enumerate all
theorems of a theory. However, such a systematic approach can
be hardly useful for some practical applications,
like looking for possible conjectures of a specific form. Our
framework allows (but does not require)
specifying partially the form of the goal: for instance, one
may specify the dominant predicate symbol in the goal atom,
or some of its arguments.

\paragraph*{Finding Deducts in Synthetic Euclidean Geometry.}

In the field of geometry, deduct candidates can also be
guessed based on an illustration, giving a concrete model.
However, these deduct candidates still have to be verified
i.e., proved. Potential deducts could also be 
listed as large disjunctions of atomic formulae,
but this method does not scale when the list of potential deducts is 
too long.

\paragraph*{Finding Deducts using Larus.}

In Larus, if the goal is given i.e., fully specified, corresponding
constraints are added to the full constraint representing a proof sought.
Let us assume that the final step of the proof is (some fixed) $n$
and, for simplicity, let us assume that the goal is just
a single atom. The corresponding constraint then includes:

\begin{enumerate}
\item \Nesting$(n) = 1$
\item \Cases$(n) = \mathit{false}$
\item \ContentsPredicate$(n,0) = $ the goal predicate symbol
\item for each argument $j$ (except for existentially quantified variables): \\
      \ContentsArgument$(n,0,j) = $ the argument from the given goal instantiated.
\end{enumerate}

\noindent
If the goal is under-specified, for instance, if the predicate symbol
is not given (it is given as \verb|_| in the TPTP file),
the third condition is just ignored. The same holds for the arguments.\footnote{Actually, underspecified arguments can be
also handled using existential quantification.}
During the solving process, if there is a model, these slots are
filled-in by some concrete values, giving a concrete goal.
Overall, support for finding under-specified deducts is very
simple. The current implementation finds one possible deduct,
but it can be extended to list all possible deducts, similarly
as for abducts (as explained in Section \ref{sec:abducts}).

\begin{example}
For Problem 4 from Section \ref{sec:intro}, Larus produces
the deduct \say {$EF \parallel GH$}. The proof
obtained with such deduct is presented in Appendix \ref{annex:varignonp4}.
\end{example}

\section{Hints and Completing Proofs}
\label{sec:hints}

Informal proofs, for instance from textbooks, are often partial and incomplete. They may even provide only a
part of a full proof, or some instructions like for filling gaps by analogy. Reconstructing proofs using such hints
is very important task, as discussed by Gowers and Hales~\cite{hales_argument_2019}:
\say{One dream was to develop an automated assistant that would
function at the level of a helpful graduate student. The senior
mathematician would suggest the main lines of the proof, and
the automated grad student would fill in the details.}


\paragraph*{Completing Proofs in Synthetic Euclidean Geometry.}
In the context of geometry, completing proofs could be interesting either
as a way to render the formalization process simpler (automation would bring in all the details that are overlooked in pen and paper proofs),
or as a tool working behind the scene
for providing guidance for what could be the next step in the proof. This objective has been studied by Richard et al.~\cite{richard_didactic_2011}. In geometry, hints can
also be based on some observations from an associated
illustration.

\paragraph*{Completing Proofs using Larus.}
Larus can be instructed to look for a proof of a given conjecture (also possibly only partially specified)
meeting some conditions (that we call \say{hints})
\cite{janicic_theorem_2022}.
Therefore, Larus can try, for instance, to reconstruct
a proof given only in outline (like proofs in textbooks).
Larus use hints in a much more general way than just
splitting the problem into sub-problems:
for instance, some hint may be used in just one proof
branch and cannot be proved itself.
Hints do not have to be ordered (one can ask for a proof
using $X$ and $Y$ in no particular order), they can be
vague, imposed only by partial constraints (\say{find a
proof that uses this particular predicate symbol},
or \say{find a proof using some specific axiom},
without the way it is instantiated, etc.).

Completing proofs in Larus is supported similarly as for abducts and under-specified goals -- by
modifying the corresponding constraints. The main
difference is that abducts and incomplete goals are
under-specified, so some constraints have to be omitted,
while partial proof introduce additional
constraints, on top of the common constraints that must
be met by all proof steps.
For expressing hints, we slightly extended the language
TPTP/fof to allow a simple but still quite expressible semantics. Some kinds of hints (not all) are illustrated below.

\begin{verbatim}
fof(hintname0, hint, r(_,_), _ , _).
fof(hintname1, hint, r(_,_), 5 , _).
fof(hintname2, hint, _, 5, ax2(_,_)).
\end{verbatim}

\noindent
The first hint specifies that some proof step will have
an atom of the form $r(\ldots, \ldots)$.
The second hint specifies that the 5th proof step will have
atom of the form $r(\ldots, \ldots)$.
The third hint specifies that in the 5th proof step the axiom
$ax2$ is applied.

\begin{example}
For Problem 5 from Section \ref{sec:intro}, Larus was able
to find a proof around 20\% faster with a suitable hint
presented in Appendix \ref{annex:varignonp5}.
\end{example}

\section{Conclusion and Future Work}

In this paper we have shown how a prover using the \say{theorem proving as constraint solving} paradigm can be
extended such that it can complete partially specified conjectures and partially specified proofs.
This extension is simple, and the implementation update is
very small. The completion algorithm is uniform, since all
three completion tasks (completing assumptions,
completing goals, completing proofs) are handled in the same spirit
-- in terms of adding or deleting some constraints.
To our knowledge, this approach is new, and we are not
aware of other systems that can address all three
sorts of completion tasks.
The presented approach is flexible as different variations
of completion tasks can be supported.
The strength of this approach is also that it can generate
both proofs that are human-readable and machine-checkable.
The proposed framework has two main limitations.
First, in current stage, it can deal only with coherent logic,
hence the theories cannot involve function symbols,
which excludes geometry proof that use (non-trivial) arithmetic.
Second, the framework cannot deal with conjectures
whose proofs are long (say, longer than 50 proof steps).

To our knowledge, there is only one other approach
in which some kind of proof is encoded, and reconstructed
from a model for the corresponding set of constraints
-- the approach in which rigid connection tableaux
are encoded as SAT and SMT instances \cite{deshane_encoding_2007,bongio_encoding_2008,mcgregor_automated_2011}. However, in this line of research, neither
machine verifiable or readable proofs, nor any of
completion tasks are considered.

The presented work can be extended in several directions.
One of our goals is to use our framework to help transfer
geometry knowledge from informal sources to proof assistants
and between proof assistants, while keeping its high-level
structure. In informal sources, statements of theorems may
be incomplete, while proofs may be given just in outline.
Still, using our approach such contents can be, at least in
some cases, completed and turned into a verifiable form.
For transferring knowledge from a proof assistant, one would
need to go into its specifics, but only to grab (some) proof
steps and make hints out of them.
We are still to explore these ideas on a larger
scale, like one geometry textbook.
In the same spirit as the work proposed by Jiang et al.~\cite{jiang_draft_2023}, our approach could be combined
with large language models to perform automatic formalization
by extracting data from natural language proofs.
More specific to abduction, we are planning to make an
in-depth comparison (both qualitative and quantitative)
of our tool to other tools for generating abducts.

\paragraph{Acknowledgement.}
The work related to this paper has been partially supported by the European Cost project CA20111 EUROProofNet.
The second author has been partially supported by the Ministry of Science
of Serbia contract 451-03-47/2023-01/200104.


\begin{thebibliography}{10}
\providecommand{\bibitemdeclare}[2]{}
\providecommand{\surnamestart}{}
\providecommand{\surnameend}{}
\providecommand{\urlprefix}{Available at }
\providecommand{\url}[1]{\texttt{#1}}
\providecommand{\href}[2]{\texttt{#2}}
\providecommand{\urlalt}[2]{\href{#1}{#2}}
\providecommand{\doi}[1]{doi:\urlalt{http://dx.doi.org/#1}{#1}}
\providecommand{\eprint}[1]{arXiv:\urlalt{https://arxiv.org/abs/#1}{#1}}
\providecommand{\bibinfo}[2]{#2}

\bibitemdeclare{book}{atocha_abductive_2006}
\bibitem{atocha_abductive_2006}
\bibinfo{author}{Aliseda \surnamestart Atocha\surnameend}
  (\bibinfo{year}{2006}): \emph{\bibinfo{title}{{ABDUCTIVE} {REASONING}}}.
\newblock {\sl \bibinfo{series}{Synthese {Library}}} \bibinfo{volume}{330},
  \bibinfo{publisher}{Kluwer Academic Publishers},
  \bibinfo{address}{Dordrecht}, \doi{10.1007/1-4020-3907-7}.

\bibitemdeclare{inproceedings}{barbosa_interactive_2023}
\bibitem{barbosa_interactive_2023}
\bibinfo{author}{Haniel \surnamestart Barbosa\surnameend},
  \bibinfo{author}{Chantal \surnamestart Keller\surnameend},
  \bibinfo{author}{Andrew \surnamestart Reynolds\surnameend},
  \bibinfo{author}{Arjun \surnamestart Viswanathan\surnameend},
  \bibinfo{author}{Cesare \surnamestart Tinelli\surnameend} \&
  \bibinfo{author}{Clark \surnamestart Barrett\surnameend}
  (\bibinfo{year}{2023}): \emph{\bibinfo{title}{An {Interactive} {SMT} {Tactic}
  in {Coq} using {Abductive} {Reasoning}}}.
\newblock In: {\sl \bibinfo{booktitle}{{EPiC} {Series} in {Computing}}},
  \bibinfo{volume}{94}, \bibinfo{publisher}{EasyChair}, pp.
  \bibinfo{pages}{11--22}, \doi{10.29007/432m}.
\newblock \bibinfo{note}{ISSN: 2398-7340}.

\bibitemdeclare{article}{beeson_proof-checking_2019}
\bibitem{beeson_proof-checking_2019}
\bibinfo{author}{Michael \surnamestart Beeson\surnameend},
  \bibinfo{author}{Julien \surnamestart Narboux\surnameend} \&
  \bibinfo{author}{Freek \surnamestart Wiedijk\surnameend}
  (\bibinfo{year}{2019}): \emph{\bibinfo{title}{Proof-checking {Euclid}}}.
\newblock {\sl \bibinfo{journal}{Annals of Mathematics and Artificial
  Intelligence}} \bibinfo{volume}{85}(\bibinfo{number}{2-4}), pp.
  \bibinfo{pages}{213--257}, \doi{10.1007/s10472-018-9606-x}.

\bibitemdeclare{inproceedings}{bezem_automating_2005}
\bibitem{bezem_automating_2005}
\bibinfo{author}{Marc \surnamestart Bezem\surnameend} \&
  \bibinfo{author}{Thierry \surnamestart Coquand\surnameend}
  (\bibinfo{year}{2005}): \emph{\bibinfo{title}{Automating {Coherent}
  {Logic}}}.
\newblock In \bibinfo{editor}{Geoff \surnamestart Sutcliffe\surnameend} \&
  \bibinfo{editor}{Andrei \surnamestart Voronkov\surnameend}, editors: {\sl
  \bibinfo{booktitle}{12th {International} {Conference} on {Logic} for
  {Programming}, {Artificial} {Intelligence}, and {Reasoning} — {LPAR}
  2005}}, {\sl \bibinfo{series}{Lecture {Notes} in {Computer} {Science}}}
  \bibinfo{volume}{3835}, \bibinfo{publisher}{Springer}, pp.
  \bibinfo{pages}{246--260}, \doi{10.1007/11591191\_18}.

\bibitemdeclare{article}{bongio_encoding_2008}
\bibitem{bongio_encoding_2008}
\bibinfo{author}{Jeremy \surnamestart Bongio\surnameend},
  \bibinfo{author}{Cyrus \surnamestart Katrak\surnameend}, \bibinfo{author}{Hai
  \surnamestart Lin\surnameend}, \bibinfo{author}{Christopher \surnamestart
  Lynch\surnameend} \& \bibinfo{author}{Ralph~Eric \surnamestart
  McGregor\surnameend} (\bibinfo{year}{2008}): \emph{\bibinfo{title}{Encoding
  {First} {Order} {Proofs} in {SMT}}}.
\newblock {\sl \bibinfo{journal}{Electron. Notes Theor. Comput. Sci.}}
  \bibinfo{volume}{198}(\bibinfo{number}{2}), pp. \bibinfo{pages}{71--84},
  \doi{10.1016/j.entcs.2008.04.081}.

\bibitemdeclare{inproceedings}{braun_tarski_2012}
\bibitem{braun_tarski_2012}
\bibinfo{author}{Gabriel \surnamestart Braun\surnameend} \&
  \bibinfo{author}{Julien \surnamestart Narboux\surnameend}
  (\bibinfo{year}{2012}): \emph{\bibinfo{title}{From {Tarski} to {Hilbert}}}.
\newblock In \bibinfo{editor}{Tetsuo \surnamestart Ida\surnameend} \&
  \bibinfo{editor}{Jacques \surnamestart Fleuriot\surnameend}, editors: {\sl
  \bibinfo{booktitle}{Post-proceedings of {Automated} {Deduction} in {Geometry}
  2012}}, {\sl \bibinfo{series}{{LNCS}}} \bibinfo{volume}{7993},
  \bibinfo{publisher}{Springer},
  pp. \bibinfo{pages}{89--109}, \doi{10.1007/978-3-642-40672-0_7}.

\bibitemdeclare{inproceedings}{chen_projection_2004}
\bibitem{chen_projection_2004}
\bibinfo{author}{XueFeng \surnamestart Chen\surnameend} \&
  \bibinfo{author}{DingKang \surnamestart Wang\surnameend}
  (\bibinfo{year}{2004}): \emph{\bibinfo{title}{The {Projection} of {Quasi}
  {Variety} and {Its} {Application} on {Geometric} {Theorem} {Proving} and
  {Formula} {Deduction}}}.
\newblock In
  {\sl \bibinfo{booktitle}{Automated {Deduction} in {Geometry}, 4th
  {International} {Workshop}, {ADG} 2002}}, {\sl \bibinfo{series}{Lecture
  {Notes} in {Computer} {Science}}} \bibinfo{volume}{2930},
  \bibinfo{publisher}{Springer}, pp. \bibinfo{pages}{21--30},
  \doi{10.1007/978-3-540-24616-9_2}.

\bibitemdeclare{inproceedings}{denecker_abduction_2002}
\bibitem{denecker_abduction_2002}
\bibinfo{author}{Marc \surnamestart Denecker\surnameend} \&
  \bibinfo{author}{Antonis~C. \surnamestart Kakas\surnameend}
  (\bibinfo{year}{2002}): \emph{\bibinfo{title}{Abduction in {Logic}
  {Programming}}}.
\newblock In {\sl
  \bibinfo{booktitle}{Computational {Logic}: {Logic} {Programming} and
  {Beyond}, {Essays} in {Honour} of {Robert} {A}. {Kowalski}, {Part} {I}}},
  {\sl \bibinfo{series}{Lecture {Notes} in {Computer} {Science}}}
  \bibinfo{volume}{2407}, \bibinfo{publisher}{Springer}, pp.
  \bibinfo{pages}{402--436}, \doi{10.1007/3-540-45628-7_16}.

\bibitemdeclare{inproceedings}{deshane_encoding_2007}
\bibitem{deshane_encoding_2007}
\bibinfo{author}{Todd \surnamestart Deshane\surnameend},
  \bibinfo{author}{Wenjin \surnamestart Hu\surnameend}, \bibinfo{author}{Patty
  \surnamestart Jablonski\surnameend}, \bibinfo{author}{Hai \surnamestart
  Lin\surnameend}, \bibinfo{author}{Christopher \surnamestart Lynch\surnameend}
  \& \bibinfo{author}{Ralph~Eric \surnamestart McGregor\surnameend}
  (\bibinfo{year}{2007}): \emph{\bibinfo{title}{Encoding {First} {Order}
  {Proofs} in {SAT}}}.
\newblock In
  {\sl \bibinfo{booktitle}{Automated {Deduction} - {CADE}-21, 21st
  {International} {Conference} on {Automated} {Deduction}}}, {\sl \bibinfo{series}{Lecture {Notes} in
  {Computer} {Science}}} \bibinfo{volume}{4603}, \bibinfo{publisher}{Springer},
  pp. \bibinfo{pages}{476--491}, \doi{10.1007/978-3-540-73595-3_35}.

\bibitemdeclare{inproceedings}{dillig_explain_2013}
\bibitem{dillig_explain_2013}
\bibinfo{author}{Isil \surnamestart Dillig\surnameend} \&
  \bibinfo{author}{Thomas \surnamestart Dillig\surnameend}
  (\bibinfo{year}{2013}): \emph{\bibinfo{title}{Explain: {A} {Tool} for
  {Performing} {Abductive} {Inference}}}.
\newblock In  {\sl
  \bibinfo{booktitle}{Computer {Aided} {Verification}}},
  \bibinfo{series}{Lecture {Notes} in {Computer} {Science}},
  \bibinfo{publisher}{Springer}, pp.
  \bibinfo{pages}{684--689}, \doi{10.1007/978-3-642-39799-8_46}.

\bibitemdeclare{incollection}{durand-guerrier_examining_2012}
\bibitem{durand-guerrier_examining_2012}
\bibinfo{author}{Viviane \surnamestart Durand-Guerrier\surnameend},
  \bibinfo{author}{Paolo \surnamestart Boero\surnameend},
  \bibinfo{author}{Nadia \surnamestart Douek\surnameend},
  \bibinfo{author}{Susanna~S. \surnamestart Epp\surnameend} \&
  \bibinfo{author}{Denis \surnamestart Tanguay\surnameend}
  (\bibinfo{year}{2012}): \emph{\bibinfo{title}{Examining the {Role} of {Logic}
  in {Teaching} {Proof}}}.
\newblock In  {\sl
  \bibinfo{booktitle}{Proof and {Proving} in {Mathematics} {Education}}}, {\sl
  \bibinfo{series}{New {ICMI} {Study} {Series}}}~\bibinfo{volume}{15},
  \bibinfo{publisher}{Springer}, pp. \bibinfo{pages}{369--389},
  \doi{10.1007/978-94-007-2129-6_16}.

\bibitemdeclare{article}{dyckhoff_geometrization_2015}
\bibitem{dyckhoff_geometrization_2015}
\bibinfo{author}{Roy \surnamestart Dyckhoff\surnameend} \&
  \bibinfo{author}{Sara \surnamestart Negri\surnameend} (\bibinfo{year}{2015}):
  \emph{\bibinfo{title}{Geometrization of first-order logic}}.
\newblock {\sl \bibinfo{journal}{The Bulletin of Symbolic Logic}}
  \bibinfo{volume}{21}, pp. \bibinfo{pages}{123--163},
  \doi{10.1017/bsl.2015.7}.

\bibitemdeclare{misc}{hales_argument_2019}
\bibitem{hales_argument_2019}
\bibinfo{author}{Thomas \surnamestart Hales\surnameend} (\bibinfo{year}{2019}):
  \emph{\bibinfo{title}{An argument for controlled natural languages in
  mathematics}}.
\newblock
  \urlprefix\url{https://jiggerwit.wordpress.com/2019/06/20/an-argument-for-controlled-natural-languages-in-mathematics/}.

\bibitemdeclare{article}{janicic_theorem_2022}
\bibitem{janicic_theorem_2022}
\bibinfo{author}{Predrag \surnamestart Janičić\surnameend} \&
  \bibinfo{author}{Julien \surnamestart Narboux\surnameend}
  (\bibinfo{year}{2022}): \emph{\bibinfo{title}{Theorem {Proving} as
  {Constraint} {Solving} with {Coherent} {Logic}}}.
\newblock {\sl \bibinfo{journal}{Journal of Automated Reasoning}}
  \bibinfo{volume}{66}(\bibinfo{number}{4}), pp. \bibinfo{pages}{689--746},
  \doi{10.1007/s10817-022-09629-z}.

\bibitemdeclare{misc}{jiang_draft_2023}
\bibitem{jiang_draft_2023}
\bibinfo{author}{Albert~Q. \surnamestart Jiang\surnameend},
  \bibinfo{author}{Sean \surnamestart Welleck\surnameend},
  \bibinfo{author}{Jin~Peng \surnamestart Zhou\surnameend},
  \bibinfo{author}{Wenda \surnamestart Li\surnameend},
  \bibinfo{author}{Jiacheng \surnamestart Liu\surnameend},
  \bibinfo{author}{Mateja \surnamestart Jamnik\surnameend},
  \bibinfo{author}{Timothée \surnamestart Lacroix\surnameend},
  \bibinfo{author}{Yuhuai \surnamestart Wu\surnameend} \&
  \bibinfo{author}{Guillaume \surnamestart Lample\surnameend}
  (\bibinfo{year}{2023}): \emph{\bibinfo{title}{Draft, {Sketch}, and {Prove}:
  {Guiding} {Formal} {Theorem} {Provers} with {Informal} {Proofs}}},
  \doi{10.48550/arXiv.2210.12283}.
\newblock \bibinfo{note}{ArXiv:2210.12283 [cs]}.

\bibitemdeclare{inproceedings}{kovacs_first-order_2013}
\bibitem{kovacs_first-order_2013}
\bibinfo{author}{Laura \surnamestart Kovács\surnameend} \&
  \bibinfo{author}{Andrei \surnamestart Voronkov\surnameend}
  (\bibinfo{year}{2013}): \emph{\bibinfo{title}{First-{Order} {Theorem}
  {Proving} and {Vampire}}}.
\newblock In {\sl
  \bibinfo{booktitle}{Computer {Aided} {Verification} - 25th {International}
  {Conference}, {CAV} 2013}}, {\sl \bibinfo{series}{Lecture {Notes} in {Computer}
  {Science}}} \bibinfo{volume}{8044}, \bibinfo{publisher}{Springer}, pp.
  \bibinfo{pages}{1--35}, \doi{10.1007/978-3-642-39799-8_1}.

\bibitemdeclare{incollection}{marquis_extending_1991}
\bibitem{marquis_extending_1991}
\bibinfo{author}{P.~\surnamestart Marquis\surnameend} (\bibinfo{year}{1991}):
  \emph{\bibinfo{title}{Extending abduction from propositional to first-order
  logic}}.
\newblock In {\sl
  \bibinfo{booktitle}{Fundamentals of {Artificial} {Intelligence} {Research}}},
  \bibinfo{publisher}{Springer},
  \doi{10.1007/3-540-54507-7_12}.

\bibitemdeclare{phdthesis}{mcgregor_automated_2011}
\bibitem{mcgregor_automated_2011}
\bibinfo{author}{Ralph~Eric \surnamestart McGregor\surnameend}
  (\bibinfo{year}{2011}): \emph{\bibinfo{title}{Automated {Theorem} {Proving}
  {Using} {SAT}}}.
\newblock \bibinfo{type}{{PhD} {Thesis}}, \bibinfo{school}{Clarkson
  University}.
\newblock
  \urlprefix\url{https://search.proquest.com/openview/b87467cab0987f591010cf19dc554fa3/1?pq-origsite=gscholar&cbl=18750&diss=y}.

\bibitemdeclare{incollection}{narboux_combining_2021}
\bibitem{narboux_combining_2021}
\bibinfo{author}{Julien \surnamestart Narboux\surnameend} \&
  \bibinfo{author}{Viviane \surnamestart Durand-Guerrier\surnameend}
  (\bibinfo{year}{2022}): \emph{\bibinfo{title}{Combining pencil/paper proofs
  and formal proofs, a challenge for Artificial Intelligence and mathematics
  education}}.
\newblock In: {\sl \bibinfo{booktitle}{Mathematics {Education} in the {Age} of
  {Artificial} {Intelligence}}}, {\sl \bibinfo{series}{{Mathematics Education
  in the Digital Era}}}~\bibinfo{volume}{17}, \bibinfo{publisher}{Springer},
  \doi{10.1007/978-3-030-86909-0_8}.

\bibitemdeclare{book}{peirce_collected_1932}
\bibitem{peirce_collected_1932}
\bibinfo{author}{Charles \surnamestart Peirce\surnameend}
  (\bibinfo{year}{1932}): \emph{\bibinfo{title}{Collected papers of {Charles}
  {Sanders} {Peirce}}}.
\newblock \bibinfo{publisher}{Belknap Press}.

\bibitemdeclare{phdthesis}{polonsky_proofs_2011}
\bibitem{polonsky_proofs_2011}
\bibinfo{author}{Andrew \surnamestart Polonsky\surnameend}
  (\bibinfo{year}{2011}): \emph{\bibinfo{title}{Proofs, {Types} and {Lambda}
  {Calculus}}}.
\newblock Ph.D.~thesis, \bibinfo{school}{University of Bergen}.

\bibitemdeclare{article}{recio_automatic_1999}
\bibitem{recio_automatic_1999}
\bibinfo{author}{T.~\surnamestart Recio\surnameend} \& \bibinfo{author}{M.~P.
  \surnamestart Vélez\surnameend} (\bibinfo{year}{1999}):
  \emph{\bibinfo{title}{Automatic {Discovery} of {Theorems} in {Elementary}
  {Geometry}}}.
\newblock {\sl \bibinfo{journal}{J. Autom. Reason.}}
  \bibinfo{volume}{23}(\bibinfo{number}{1}), pp. \bibinfo{pages}{63--82},
  \doi{10.1023/A:1006135322108}.

\bibitemdeclare{inproceedings}{reynolds_scalable_2020}
\bibitem{reynolds_scalable_2020}
\bibinfo{author}{Andrew \surnamestart Reynolds\surnameend},
  \bibinfo{author}{Haniel \surnamestart Barbosa\surnameend},
  \bibinfo{author}{Daniel \surnamestart Larraz\surnameend} \&
  \bibinfo{author}{Cesare \surnamestart Tinelli\surnameend}
  (\bibinfo{year}{2020}): \emph{\bibinfo{title}{Scalable {Algorithms} for
  {Abduction} via {Enumerative} {Syntax}-{Guided} {Synthesis}}}.
\newblock In  {\sl \bibinfo{booktitle}{Automated {Reasoning} - 10th
  {International} {Joint} {Conference}, {IJCAR} 2020, {Part} {I}}}, {\sl \bibinfo{series}{Lecture {Notes}
  in {Computer} {Science}}} \bibinfo{volume}{12166},
  \bibinfo{publisher}{Springer}, pp. \bibinfo{pages}{141--160},
  \doi{10.1007/978-3-030-51074-9_9}.

\bibitemdeclare{article}{richard_didactic_2011}
\bibitem{richard_didactic_2011}
\bibinfo{author}{Philippe~R. \surnamestart Richard\surnameend},
  \bibinfo{author}{Josep~Maria \surnamestart Fortuny\surnameend},
  \bibinfo{author}{Michel \surnamestart Gagnon\surnameend},
  \bibinfo{author}{Nicolas \surnamestart Leduc\surnameend},
  \bibinfo{author}{Eloi \surnamestart Puertas\surnameend} \&
  \bibinfo{author}{Michèle \surnamestart Tessier-Baillargeon\surnameend}
  (\bibinfo{year}{2011}): \emph{\bibinfo{title}{Didactic and theoretical-based
  perspectives in the experimental development of an intelligent tutorial
  system for the learning of geometry}}.
\newblock {\sl \bibinfo{journal}{ZDM}}
  \bibinfo{volume}{43}(\bibinfo{number}{3}), pp. \bibinfo{pages}{425--439},
  \doi{10.1007/s11858-011-0320-y}.

\bibitemdeclare{article}{russo_use_2001}
\bibitem{russo_use_2001}
\bibinfo{author}{Alessandra \surnamestart Russo\surnameend} \&
  \bibinfo{author}{Bashar \surnamestart Nuseibeh\surnameend}
  (\bibinfo{year}{2001}): \emph{\bibinfo{title}{On {The} {Use} {Of} {Logical} {Abduction} {In} {Software} {Engineering}}}.
  In {\sl {Handbook of Software Engineering and Knowledge Engineering}},
\newblock \doi{10.1142/9789812389718_0037}.

\bibitemdeclare{book}{schwabhauser_metamathematische_1983}
\bibitem{schwabhauser_metamathematische_1983}
\bibinfo{author}{Wolfram \surnamestart Schwabhäuser\surnameend},
  \bibinfo{author}{Wanda \surnamestart Szmielew\surnameend} \&
  \bibinfo{author}{Alfred \surnamestart Tarski\surnameend}
  (\bibinfo{year}{1983}): \emph{\bibinfo{title}{Metamathematische {Methoden} in
  der {Geometrie}}}.
\newblock \bibinfo{publisher}{Springer}.
\newblock \doi{10.1007/978-3-642-69418-9}.

\bibitemdeclare{incollection}{stojanovic_vernacular_2014}
\bibitem{stojanovic_vernacular_2014}
\bibinfo{author}{Sana \surnamestart Stojanović\surnameend},
  \bibinfo{author}{Julien \surnamestart Narboux\surnameend},
  \bibinfo{author}{Marc \surnamestart Bezem\surnameend} \&
  \bibinfo{author}{Predrag \surnamestart Janičić\surnameend}
  (\bibinfo{year}{2014}): \emph{\bibinfo{title}{A {Vernacular} for {Coherent}
  {Logic}}}.
\newblock In {\sl
  \bibinfo{booktitle}{Intelligent {Computer} {Mathematics}}}, {\sl
  \bibinfo{series}{Lecture {Notes} in {Computer} {Science}}}
  \bibinfo{volume}{8543}, \bibinfo{publisher}{Springer}, pp. \bibinfo{pages}{388--403},
  \doi{10.1007/978-3-319-08434-3_28}.

\bibitemdeclare{article}{wu_decision_1978}
\bibitem{wu_decision_1978}
\bibinfo{author}{Wen-Tsun \surnamestart Wu\surnameend} (\bibinfo{year}{1978}):
  \emph{\bibinfo{title}{On the {Decision} {Problem} and the {Mechanization} of
  {Theorem}-{Proving} in {Elementary} {Geometry}}}.
\newblock \bibinfo{volume}{21}, \bibinfo{publisher}{Scientia Sinica}, pp.
  \bibinfo{pages}{157--179}.

\end{thebibliography}


\section{Appendix}

In this appendix, we provide a complete list of lemmas
and axioms (in coherent logic form) used in our examples,
and the results obtained using Larus. The results were
obtained on a PC computer with Intel(R) Core(TM) i7-8565U CPU @ 1.80GHz
processor running under Linux
(the time spent should give just a general picture of
the efficiency of the system).

\subsection{Problem 1: Varignon's Theorem}
\label{annex:varignonp1}

The TPTP file used for Problem 1 is the following:

\lstset{
  basicstyle=\ttfamily,
  columns=fullflexible,
  keepspaces=true,
}

{\footnotesize
\begin{tcolorbox}[breakable]
\begin{lstlisting}[breaklines]
fof(triangle_mid_par_strict, axiom, (! [A, B, C, P, Q] : ( ((~ col(A,B,C)) & midpoint(B,P,C) & midpoint(A,Q,C)) => par(A,B,Q,P)))).
fof(lemma_par_trans, axiom, (! [A, B, C, D, E, F] : ((par(A,B,C,D) & par(C,D,E,F) & (~col(A,B,E))) => par(A,B,E,F)))).
fof(defparallelogram2,axiom, (! [A,B,C,D] : ((par(A,B,C,D) & par(A,D,B,C)) => ((pG(A,B,C,D)))))).
fof(lemma_parallelNC,axiom, (! [A,B,C,D] : ((par(A,B,C,D)) => ((~ (col(A,B,C)) & ~ (col(A,C,D)) & ~ (col(B,C,D)) & ~ (col(A,B,D))))))).
fof(lemma_parallelflip,axiom, (! [A,B,C,D] : ((par(A,B,C,D)) => ((par(B,A,C,D) & par(A,B,D,C) & par(B,A,D,C)))))).
fof(lemma_parallelsymmetric,axiom, (! [A,B,C,D] : ((par(A,B,C,D)) => ((par(C,D,A,B)))))).
fof(midpoint_sym, axiom, (! [A, B, I] : (midpoint(A,I,B) => midpoint(B,I,A)))).
fof(lemma_tP_trans, axiom,  (! [A, B, C, D, E, F] : ((tP(A,B,C,D) & tP(C,D,E,F)) => tP(A,B,E,F)))).

fof(th_varignon,conjecture,(! [A,B,C,D,E,F,G,H] : (( (~(col(B,D,A))) & (~(col(B,D,C))) & (~(col(A,C,B))) & (~(col(A,C,D))) & (~ (col(E,F,G))) & midpoint(A,E,B) & midpoint(B,F,C) & midpoint(C,G,D) & midpoint(A,H,D)) => pG(E,F,G,H) ))).
\end{lstlisting}
\end{tcolorbox}
}

If Larus is invoked as:
\verb|./larus -l100 -m8|
(\verb|-l100| means the time limit is 100s, \verb|-m8|
means that we look for a proof with 8 or fewer steps),
it produces the following proof in 2s:

\setcounter{proofstepnum}{0}

\begin{tcolorbox}[breakable]
\noindent Consider arbitrary $a$, $b$, $c$, $d$, $e$, $f$, $g$, $h$ such that: \begin{itemize} 

\item  $\neg col(b, d, a)$, 
\item  $\neg col(b, d, c)$, 
\item  $\neg col(a, c, b)$, 
\item  $\neg col(a, c, d)$, 
\item  $\neg col(e, f, g)$, 
\item  $b \neq d$, 
\item  $a \neq c$, 
\item  $midpoint(a, e, b)$, 
\item  $midpoint(b, f, c)$, 
\item  $midpoint(c, g, d)$, 
\item  $midpoint(a, h, d)$. 
\end{itemize} 
It should be proved that $pG(e, f, g, h)$.

\proofstep{0}{$par(a, c, h, g)$ ({\scriptsize by MP, from $\neg col(a, c, d)$, $midpoint(c, g, d)$, $midpoint(a, h, d)$ using axiom triangle\_mid\_par\_strict; instantiation:  $A$ $\mapsto$  $a$,  $B$ $\mapsto$  $c$,  $C$ $\mapsto$  $d$,  $P$ $\mapsto$  $g$,  $Q$ $\mapsto$  $h$}) }
\proofstep{0}{$par(b, d, f, g)$ ({\scriptsize by MP, from $\neg col(b, d, c)$, $midpoint(c, g, d)$, $midpoint(b, f, c)$ using axiom triangle\_mid\_par\_strict; instantiation:  $A$ $\mapsto$  $b$,  $B$ $\mapsto$  $d$,  $C$ $\mapsto$  $c$,  $P$ $\mapsto$  $g$,  $Q$ $\mapsto$  $f$}) }
\proofstep{0}{$par(a, c, e, f)$ ({\scriptsize by MP, from $\neg col(a, c, b)$, $midpoint(b, f, c)$, $midpoint(a, e, b)$ using axiom triangle\_mid\_par\_strict; instantiation:  $A$ $\mapsto$  $a$,  $B$ $\mapsto$  $c$,  $C$ $\mapsto$  $b$,  $P$ $\mapsto$  $f$,  $Q$ $\mapsto$  $e$}) }
\proofstep{0}{$par(b, d, e, h)$ ({\scriptsize by MP, from $\neg col(b, d, a)$, $midpoint(a, h, d)$, $midpoint(a, e, b)$ using axiom triangle\_mid\_par\_strict; instantiation:  $A$ $\mapsto$  $b$,  $B$ $\mapsto$  $d$,  $C$ $\mapsto$  $a$,  $P$ $\mapsto$  $h$,  $Q$ $\mapsto$  $e$}) }
\proofstep{0}{$par(e, f, g, h)$ ({\scriptsize by MP, from $par(a, c, e, f)$, $par(a, c, h, g)$, $\neg col(e, f, g)$ using axiom lemma\_par\_trans; instantiation:  $A$ $\mapsto$  $e$,  $B$ $\mapsto$  $f$,  $C$ $\mapsto$  $a$,  $D$ $\mapsto$  $c$,  $E$ $\mapsto$  $g$,  $F$ $\mapsto$  $h$}) }
\proofstep{0}{$par(f, g, h, e)$ ({\scriptsize by MP, from $par(b, d, f, g)$, $par(b, d, e, h)$, $par(e, f, g, h)$ using axiom lemma\_par\_trans; instantiation:  $A$ $\mapsto$  $f$,  $B$ $\mapsto$  $g$,  $C$ $\mapsto$  $d$,  $D$ $\mapsto$  $b$,  $E$ $\mapsto$  $h$,  $F$ $\mapsto$  $e$}) }
\proofstep{0}{$pG(e, f, g, h)$ ({\scriptsize by MP, from $par(e, f, g, h)$, $par(f, g, h, e)$ using axiom defparallelogram2; instantiation:  $A$ $\mapsto$  $e$,  $B$ $\mapsto$  $f$,  $C$ $\mapsto$  $g$,  $D$ $\mapsto$  $h$}) }
\proofstep{0}{Proved by assumption! ({\scriptsize by QEDas})}
\end{tcolorbox}

\subsection{Problem 2: First Inverse Problem}
\label{annex:varignonp2}

The list of axioms used for the first inverse problem (Problem 2) is the same as in Section \ref{annex:varignonp1}.
Only the conjecture is different --
the assumption \verb|midpoint(A,H,D)| is ommitted:

{\footnotesize
\begin{tcolorbox}[breakable]
\begin{lstlisting}[breaklines]
fof(th_varignon,conjecture,(! [A,B,C,D,E,F,G,H] : (( (~(col(B,D,A))) & (~(col(B,D,C))) & (~(col(A,C,B))) & (~(col(A,C,D))) & (~ (col(E,F,G))) & (B != D) & (A != C) & midpoint(A,E,B) & midpoint(B,F,C) & midpoint(C,G,D)) => pG(E,F,G,H) ))).
\end{lstlisting}
\end{tcolorbox}
}

If Larus is invoked as:
\verb|./larus -l100 -m8 -b1|
(\verb|-l100| means the time limit is 100s, \verb|-m8|
means that we look for a proof with 8 or fewer steps,
\verb|-b1| means that we look for one atomic formula
as an abduct), it finds a first consistent abduct (after two
inconsistent ones) and produces the following human-readable
proof in 3.26 seconds
(the abduct found is highlighted):

\setcounter{proofstepnum}{0}

\begin{tcolorbox}[breakable]
\noindent Consider arbitrary $a$, $b$, $c$, $d$, $e$, $f$, $g$, $h$ such that: 

\begin{itemize} 
\item  $\neg col(b, d, a)$, 
\item  $\neg col(b, d, c)$, 
\item  $\neg col(a, c, b)$, 
\item  $\neg col(a, c, d)$, 
\item  $\neg col(e, f, g)$, 
\item  $b \neq d$, 
\item  $a \neq c$, 
\item  $midpoint(a, e, b)$, 
\item  $midpoint(b, f, c)$, 
\item  $midpoint(c, g, d)$. 
\end{itemize} 
It should be proved that $pG(e, f, g, h)$.

\definecolor{shadecolor}{RGB}{220,220,120}
\colorbox{shadecolor}{
\begin{minipage}{0.96\textwidth}
Abducts found:
\begin{itemize}
\item $midpoint(d, h, a)$
\end{itemize}
\end{minipage}
}

\proofstep{0}{$par(a, c, e, f)$ ({\scriptsize by MP, from $\neg col(a, c, b)$, $midpoint(b, f, c)$, $midpoint(a, e, b)$ using axiom triangle\_mid\_par\_strict; instantiation:  $A$ $\mapsto$  $a$,  $B$ $\mapsto$  $c$,  $C$ $\mapsto$  $b$,  $P$ $\mapsto$  $f$,  $Q$ $\mapsto$  $e$}) }
\proofstep{0}{$par(b, d, f, g)$ ({\scriptsize by MP, from $\neg col(b, d, c)$, $midpoint(c, g, d)$, $midpoint(b, f, c)$ using axiom triangle\_mid\_par\_strict; instantiation:  $A$ $\mapsto$  $b$,  $B$ $\mapsto$  $d$,  $C$ $\mapsto$  $c$,  $P$ $\mapsto$  $g$,  $Q$ $\mapsto$  $f$}) }
\proofstep{0}{$par(b, d, e, h)$ ({\scriptsize by MP, from $\neg col(b, d, a)$, $midpoint(d, h, a)$, $midpoint(a, e, b)$ using axiom triangle\_mid\_par\_strict; instantiation:  $A$ $\mapsto$  $b$,  $B$ $\mapsto$  $d$,  $C$ $\mapsto$  $a$,  $P$ $\mapsto$  $h$,  $Q$ $\mapsto$  $e$}) }
\proofstep{0}{$par(a, c, h, g)$ ({\scriptsize by MP, from $\neg col(a, c, d)$, $midpoint(c, g, d)$, $midpoint(d, h, a)$ using axiom triangle\_mid\_par\_strict; instantiation:  $A$ $\mapsto$  $a$,  $B$ $\mapsto$  $c$,  $C$ $\mapsto$  $d$,  $P$ $\mapsto$  $g$,  $Q$ $\mapsto$  $h$}) }
\proofstep{0}{$par(e, f, g, h)$ ({\scriptsize by MP, from $par(a, c, e, f)$, $par(a, c, h, g)$, $\neg col(e, f, g)$ using axiom lemma\_par\_trans; instantiation:  $A$ $\mapsto$  $e$,  $B$ $\mapsto$  $f$,  $C$ $\mapsto$  $a$,  $D$ $\mapsto$  $c$,  $E$ $\mapsto$  $g$,  $F$ $\mapsto$  $h$}) }
\proofstep{0}{$par(e, h, g, f)$ ({\scriptsize by MP, from $par(b, d, e, h)$, $par(b, d, f, g)$, $par(e, f, g, h)$ using axiom lemma\_par\_trans; instantiation:  $A$ $\mapsto$  $e$,  $B$ $\mapsto$  $h$,  $C$ $\mapsto$  $b$,  $D$ $\mapsto$  $d$,  $E$ $\mapsto$  $g$,  $F$ $\mapsto$  $f$}) }
\proofstep{0}{$pG(e, f, g, h)$ ({\scriptsize by MP, from $par(e, f, g, h)$, $par(e, h, g, f)$ using axiom defparallelogram2; instantiation:  $A$ $\mapsto$  $e$,  $B$ $\mapsto$  $f$,  $C$ $\mapsto$  $g$,  $D$ $\mapsto$  $h$}) }
\proofstep{0}{Proved by assumption! ({\scriptsize by QEDas})}
\end{tcolorbox}

\subsection{Problem 3: Second Inverse Problem}
\label{annex:varignonp3}

The list of axioms used for the second inverse problem (Problem 3) is the same as in section \ref{annex:varignonp1}, extended with the following axioms.

{\footnotesize
\begin{tcolorbox}[breakable]
\begin{lstlisting}[breaklines]
fof(defmidpoint,axiom, (! [A,B,C] : ((midpoint(A,B,C)) => ((betS(A,B,C) & cong(A,B,B,C)))))).
fof(defmidpoint2,axiom, (! [A,B,C] : ((betS(A,B,C) & cong(A,B,B,C)) => ((midpoint(A,B,C)))))).
fof(midpoint_NC, axiom, (! [A, B, I] : ((midpoint(A,I,B) & (A != B)) => ( (A != I) & ( B != I))))).
fof(defrectangle,axiom, (! [A,B,C,D] : ((rectangle(A,B,C,D)) => ((pG(A,B,C,D) & per(A,B,C) & per(B,C,D) & per(C,D,A) & per(D,A,B)))))).
fof(defrectangle2a,axiom, (! [A,B,C,D] : ((pG(A,B,C,D) & per(A,B,C)) => rectangle(A,B,C,D)))).
fof(defrectangle2b,axiom, (! [A,B,C,D] : ((pG(A,B,C,D) & per(B,C,D)) => rectangle(A,B,C,D)))).
fof(defrectangle2c,axiom, (! [A,B,C,D] : ((pG(A,B,C,D) & per(C,D,A)) => rectangle(A,B,C,D)))).
fof(defrectangle2d,axiom, (! [A,B,C,D] : ((pG(A,B,C,D) & per(D,A,B)) => rectangle(A,B,C,D)))).
fof(defrectangle2e,axiom, (! [A,B,C,D] : ((per(A,B,C) & per(B,C,D) & per(C,D,A) & per(D,A,B)) => rectangle(A,B,C,D)))).
%fof(defrectangle3a,axiom, (! [A,B,C,D] : (? [X] : ((rectangle(A,B,C,D)) => cong(A,C,B,D) & midpoint(A,X,C) & midpoint(B,X,D))))).
fof(defrectangle3b,axiom, (! [A,B,C,D,X] : ((cong(A,C,B,D) & midpoint(A,X,C) & midpoint(B,X,D)) => rectangle(A,B,C,D)))).
fof(defrectangle4a,axiom, (! [A,B,C,D] : ((rectangle(A,B,C,D)) => (pG(A,B,C,D) & cong(A,C,B,D))))).
fof(defrectangle4b,axiom, (! [A,B,C,D] : ((pG(A,B,C,D) & cong(A,C,B,D)) => rectangle(A,B,C,D)))).
fof(lemma_8_2,axiom, (! [A,B,C] : ((per(A,B,C)) => ((per(C,B,A)))))).
fof(varignon_th,axiom,(! [A,B,C,D,E,F,G,H] : (( (~(col(B,D,A))) & (~(col(B,D,C))) & (~(col(A,C,B))) & (~(col(A,C,D))) & (~ (col(G,F,E))) & (B != D) & (A != C) & midpoint(A,E,B) & midpoint(B,F,C) & midpoint(C,G,D) & midpoint(A,H,D)) => pG(E,F,G,H) ))).
\end{lstlisting}
\end{tcolorbox}
}

The conjecture is also different -- the goal is to find under
which assumption the quadrilateral $EFGH$ is a rectangle.

{\footnotesize
\begin{tcolorbox}[breakable]
\begin{lstlisting}[breaklines]
fof(th_varignon_rect,conjecture,(! [A,B,C,D,E,F,G,H] : (( (~(col(B,D,A))) & (~(col(B,D,C))) & (~(col(A,C,B))) & (~(col(A,C,D))) & (~ (col(G,F,E))) & (B != D) & (A != C) & midpoint(A,E,B) & midpoint(B,F,C) & midpoint(C,G,D) & midpoint(A,H,D)) => rectangle(E,F,G,H) ))).
\end{lstlisting}
\end{tcolorbox}
}

If Larus is invoked as: \verb|./larus -l100 -m8 -b1|,
it produces the following human-readable proof in 14s
(the abduct found is highlighted):

\setcounter{proofstepnum}{0}

\begin{tcolorbox}[breakable]
\setcounter{proofstepnum}{0}

\noindent
Consider arbitrary $a$, $b$, $c$, $d$, $e$, $f$, $g$, $h$ such that: \begin{itemize}

\item  $\neg col(b, d, a)$, 
\item  $\neg col(b, d, c)$, 
\item  $\neg col(a, c, b)$, 
\item  $\neg col(a, c, d)$, 
\item  $\neg col(f, g, e)$, 
\item  $b \neq d$, 
\item  $a \neq c$, 
\item  $midpoint(a, e, b)$, 
\item  $midpoint(b, f, c)$, 
\item  $midpoint(c, g, d)$, 
\item  $midpoint(a, h, d)$. 
\end{itemize} 
It should be proved that $rectangle(e, f, g, h)$.

\definecolor{shadecolor}{RGB}{220,220,120}
\colorbox{shadecolor}{
\begin{minipage}{0.96\textwidth}
Abducts found: 
\begin{itemize} 
\item $cong(e, g, h, f)$
\end{itemize} 
\end{minipage}}

\proofstep{0}{$midpoint(b, e, a)$ ({\scriptsize by MP, from $midpoint(a, e, b)$, $midpoint(a, e, b)$ using axiom defmidpoint2; instantiation:  $A$ $\mapsto$  $b$,  $B$ $\mapsto$  $e$,  $C$ $\mapsto$  $a$}) }
\proofstep{0}{$pG(e, f, g, h)$ ({\scriptsize by MP, from $\neg col(b, d, a)$, $\neg col(b, d, c)$, $\neg col(a, c, b)$, $\neg col(a, c, d)$, $\neg col(f, g, e)$, $b \neq d$, $a \neq c$, $midpoint(b, e, a)$, $midpoint(b, f, c)$, $midpoint(c, g, d)$, $midpoint(a, h, d)$ using axiom varignon\_th; instantiation:  $A$ $\mapsto$  $a$,  $B$ $\mapsto$  $b$,  $C$ $\mapsto$  $c$,  $D$ $\mapsto$  $d$,  $I$ $\mapsto$  $e$,  $J$ $\mapsto$  $f$,  $K$ $\mapsto$  $g$,  $L$ $\mapsto$  $h$}) }
\proofstep{0}{$rectangle(e, f, g, h)$ ({\scriptsize by MP, from $pG(e, f, g, h)$, $cong(e, g, h, f)$ using axiom defrectangle4b; instantiation:  $A$ $\mapsto$  $e$,  $B$ $\mapsto$  $f$,  $C$ $\mapsto$  $g$,  $D$ $\mapsto$  $h$}) }
\proofstep{0}{$rectangle(e, f, g, h)$ ({\scriptsize by MP, from $rectangle(e, f, g, h)$, $rectangle(e, f, g, h)$, $rectangle(e, f, g, h)$, $rectangle(e, f, g, h)$ using axiom defrectangle2e; instantiation:  $A$ $\mapsto$  $e$,  $B$ $\mapsto$  $f$,  $C$ $\mapsto$  $g$,  $D$ $\mapsto$  $h$}) }
\proofstep{0}{Proved by assumption! ({\scriptsize by QEDas})}
\end{tcolorbox}

\subsection{Problem 4: Partially Specified Goal}
\label{annex:varignonp4}

The list of axioms used for Problem 4 is the same as presented
in Section \ref{annex:varignonp1}. Only the conjecture is
different: the goal does not have the predicate symbol specified:

{\footnotesize
\begin{tcolorbox}[breakable]
\begin{lstlisting}[breaklines]
fof(th_varignon,conjecture,(! [A,B,C,D,E,F,G,H] : (( (~(col(B,D,A))) & (~(col(B,D,C))) & (~(col(A,C,B))) & (~(col(A,C,D))) & (~ (col(E,F,G))) & (B != D) & (A != C) & midpoint(A,E,B) & midpoint(B,F,C) & midpoint(C,G,D) & midpoint(A,H,D)) => _(E,F,G,H) ))).
\end{lstlisting}
\end{tcolorbox}
}

\noindent
If Larus is invoked as
\verb|./larus -l100 -m8|,
it produces the following human-readable proof
(for the goal $par(e, f, g, h)$, highlighted in the proof)
in 2s:

\begin{tcolorbox}[breakable]
\setcounter{proofstepnum}{0}

\noindent Consider arbitrary $a$, $b$, $c$, $d$, $e$, $f$, $g$, $h$ such that: \begin{itemize}

\item  $\neg col(b, d, a)$,
\item  $\neg col(b, d, c)$,
\item  $\neg col(a, c, b)$,
\item  $\neg col(a, c, d)$,
\item  $\neg col(e, f, g)$,
\item  $b \neq d$,
\item  $a \neq c$,
\item  $midpoint(a, e, b)$,
\item  $midpoint(b, f, c)$,
\item  $midpoint(c, g, d)$,
\item  $midpoint(a, h, d)$.
\end{itemize}

\definecolor{shadecolor}{RGB}{220,220,120}
\colorbox{shadecolor}{
\begin{minipage}{0.96\textwidth}
It should be proved that $\_(e, f, g, h)$.
\end{minipage}}

\vspace{5pt}

\proofstep{0}{$par(a, c, e, f)$ ({\scriptsize by MP, from $\neg col(a, c, b)$, $midpoint(b, f, c)$, $midpoint(a, e, b)$ using axiom triangle\_mid\_par\_strict; instantiation:  $A$ $\mapsto$  $a$,  $B$ $\mapsto$  $c$,  $C$ $\mapsto$  $b$,  $P$ $\mapsto$  $f$,  $Q$ $\mapsto$  $e$}) }
\proofstep{0}{$par(a, c, h, g)$ ({\scriptsize by MP, from $\neg col(a, c, d)$, $midpoint(c, g, d)$, $midpoint(a, h, d)$ using axiom triangle\_mid\_par\_strict; instantiation:  $A$ $\mapsto$  $a$,  $B$ $\mapsto$  $c$,  $C$ $\mapsto$  $d$,  $P$ $\mapsto$  $g$,  $Q$ $\mapsto$  $h$}) }
\proofstep{0}{
\definecolor{shadecolor}{RGB}{220,220,120}
\colorbox{shadecolor}{
\begin{minipage}{0.18\textwidth}
$par(e, f, g, h)$
\end{minipage}}
({\scriptsize by MP, from $par(a, c, e, f)$, $par(a, c, h, g)$, $\neg col(e, f, g)$ using axiom lemma\_par\_trans; instantiation:  $A$ $\mapsto$  $e$,  $B$ $\mapsto$  $f$,  $C$ $\mapsto$  $a$,  $D$ $\mapsto$  $c$,  $E$ $\mapsto$  $g$,  $F$ $\mapsto$  $h$}) }
\proofstep{0}{Proved by assumption! ({\scriptsize by QEDas})}
\end{tcolorbox}

\subsection{Problem 5: Partially Specified Proof}
\label{annex:varignonp5}

The list of axioms used for Problem 5 is
as presented in
Section \ref{annex:varignonp3} (with the axiom \verb|defrectangle3a| deleted). The conjecture is the same plus the abduct as an assumption,
but we add the following hint:

{\footnotesize
\begin{tcolorbox}[breakable]
\begin{Verbatim}[breaklines=true]
fof(hint1,hint,_,_,defrectangle4b(4,5,6,7)).
\end{Verbatim}
\end{tcolorbox}
}

\noindent
If Larus is invoked as \verb|./larus -l100 -m8|,
it produces the same proof as in Section \ref{annex:varignonp3}
in 4s, while if the hint is omitted, it takes 5s.

\end{document}